\documentclass{article}

\usepackage{microtype}
\usepackage{graphicx}
\usepackage{subfigure}
\usepackage{booktabs} % for professional tables
\usepackage{amsmath}
\usepackage{amsfonts}
\usepackage{mathtools}
\usepackage{algorithm}
\usepackage{algorithmic}

\usepackage{hyperref}

\usepackage[accepted]{icml2020}

\icmltitlerunning{True to the Model or True to the Data?}

\begin{document}

\twocolumn[
\icmltitle{True to the Model or True to the Data?}



\icmlsetsymbol{equal}{*}

\begin{icmlauthorlist}
\icmlauthor{Hugh Chen}{equal,uw}
\icmlauthor{Joseph D. Janizek}{equal,uw}
\icmlauthor{Scott Lundberg}{msr}
\icmlauthor{Su-In Lee}{uw}
\end{icmlauthorlist}

\icmlaffiliation{uw}{University of Washington}
\icmlaffiliation{msr}{Microsoft Research}
% \icmlaffiliation{ed}{School of Computation, University of Edenborrow, Edenborrow, United Kingdom}

\icmlcorrespondingauthor{Su-In Lee}{suinlee@cs.washington.edu}
% \icmlcorrespondingauthor{Eee Pppp}{ep@eden.co.uk}

\icmlkeywords{Machine Learning, ICML}

\vskip 0.3in
]



\printAffiliationsAndNotice{\icmlEqualContribution} % otherwise use the standard text.

\begin{abstract}
A variety of recent papers discuss the application of Shapley values, a concept for explaining coalitional games, for feature attribution in machine learning.  However, the correct way to connect a machine learning model to a coalitional game has been a source of controversy. The two main approaches that have been proposed differ in the way that they condition on known features, using either (1) an interventional or (2) an observational conditional expectation. While previous work has argued that one of the two approaches is preferable \textit{in general}, we argue that the choice is \textit{application dependent}. Furthermore, we argue that the choice comes down to whether it is desirable to be \textit{true to the model} or \textit{true to the data}. We use linear models to investigate this choice. After deriving an efficient method for calculating observational conditional expectation Shapley values for linear models\footnote{\url{https://github.com/slundberg/shap/blob/master/shap/explainers/linear.py}}, we investigate how correlation in simulated data impacts the convergence of observational conditional expectation Shapley values. Finally, we present two real data examples that we consider to be representative of possible use cases for feature attribution -- (1) credit risk modeling and (2) biological discovery. We show how a different choice of value function performs better in each scenario, and how possible attributions are impacted by modeling choices.
\end{abstract}

\section{Shapley values}

One of the most popular approaches to machine learning interpretability in recent years has involved using Shapley values to attribute importance to features \cite{vstrumbelj2014explaining,lundberg2017unified}. The Shapley value is a concept from coalitional game theory that fairly allocates the surplus generated by the grand coalition in a game to each of its players \cite{shapley1953value}. In this general sense, the Shapley value allocated to a player $i$ is defined as:

\begin{align}
\phi_i = \frac{1}{M!} \sum_R [ v(S^R\cup i) - v(S^R)],
\end{align}

where $R$ is one possible permutation of the order in which the players join the coalition, $S^R$ is the set of players  joining the coalition \textit{before} player $i$, and $v:S\in \mathcal{P}(M)\to\mathbb{R}^1$ is a coalitional game that maps from the power set $\mathcal{P}$ of all players $M$ to a scalar value.

% \subsection{Shapley axioms in general}

% \hugh{May be able to cut this for the sake of brevity (since it's a workshop paper)}

\subsection{Choice of value function}

While the Shapley value is provably the unique solution which satisfies a variety of axioms for an abstract n-person game, figuring out how to represent a machine learning model ($f$) as a coalitional game ($v$) is non-trivial. Previous work has suggested a variety of different functional forms for $v$, for tasks like data valuation and global feature importance \cite{ghorbani2019data,covert2020understanding}. In this work, we focus on local feature attribution -- trying to understand how much each feature contributed to the output of a model for a particular sample. For this application, the reward of the game is typically the conditional expectation of the model's output, where the players in the game are the known features in the conditional expectation. There are two ways the model's output ($f:x\in\mathbb{R}^{|N|\times 1}\to\mathbb{R}^1$) for a particular sample is used to define $v(S)$:

\textit{1. Observational conditional expectation:} This is the formulation in \cite{lundberg2017unified,aas2019explaining,frye2019asymmetric}.  The coalitional game is 
\begin{equation}
v(S)=\mathbb{E}[f(x)| S]    
\end{equation}
where conditioning on $S$ means considering the input $X$ to be a random variable where the features in $S$ are known ($\mathbb{E}[f(X)|X_S=x_S]$).

\textit{2. Interventional conditional expectation:} This approach is endorsed by \cite{janzing2019feature,sundararajan2019many, datta2016algorithmic}, and in practice is used to approximate the observational conditional expectation in \cite{lundberg2017unified}. Here the coalitional game is defined as:
\begin{equation}
v(S)=\mathbb{E}[f(x)|do(S)]
\end{equation}
where we ``intervene'' on the features by breaking the dependence between features in $S$ and the remaining features. We refer to Shapley values obtained with either approach as either observational or interventional Shapley values.  

\subsection{Apparent problems with choice of value function}

Previous work has pointed out issues with each choice of value function. For example, \citet{janzing2019feature} and \citet{sundararajan2019many} both point out that the observational Shapley value can attribute importance to \textit{irrelevant features} -- features which were not used by the model. While this does not violate the original Shapley axioms, it does violate a new axiom called \textit{Dummy} proposed by \citet{sundararajan2019many}, which requires that a feature $i$ will get attribution $\phi_i = 0$, if for any two values $x_i$ and $x_i'$ and for every value $x_{N \setminus i}$, $f(x_i; x_{N \setminus i}) = f(x_i'; x_{N \setminus i})$. On the other hand, papers like \citet{frye2019asymmetric} have noted that using the interventional Shapley value (which breaks the dependence between features) will lead to evaluating the model on ``impossible data points'' that lie off the true data manifold.

While recent work has gone so far as to suggest that having two separate approaches presents an irreconcilable problem with using Shapley values for feature attribution \cite{kumar2020problems}, in this paper, we argue that rather than representing some critical flaw in using the Shapley value for feature attribution, each approach is meaningful when applied in the proper context. Further, we argue that this choice depends on whether you want attributions that reflect the behavior of a particular model, or attributions that reflect the correlations in the data.

\section{Linear SHAP}

In order to understand both approaches, we will focus on \textit{linear models} where we present a novel algorithm to compute the observational Shapley values.  Moving forward, $f(x) = \beta x + b$ where $\beta\in \mathbb{R}^{1\times |N|}$ is a row vector and $b\in \mathbb{R}^1$ a scalar.  

\subsection{Interventional conditional expectation}

For an interventional conditional expectation, the Shapley values (which we denote as $\phi_i(f,x)$) are:
\begin{align}
\phi_i(f,x)=\beta_i (x_i - \mu_i)
\end{align}
This was shown for independent features \cite{aas2019explaining} and the interventional conditional expectation gives the same explanations.

\subsection{Observational conditional expectation}

Computing the Shapley values for an observational conditional expectation is substantially harder, with a number of proposed algorithms for doing so. \citet{sundararajan2019many} utilizes the empirical distribution, which often assigns zero probability to plausible samples even for large samples.  \citet{mase2019explaining} extends this empirical distribution by including a similarity metric.  In \citet{aas2019explaining}, the unknown features are sampled from either a multivariate gaussian conditional, a gaussian copula conditional, or an empirical conditional distribution. In \citet{frye2019asymmetric}, the conditional is modeled using an autoencoder. For a linear model, the problem reduces to estimating the conditional expectation of $x$ given different subsets\footnote{The key observation is the for a linear $f(x)$, the expectation (and the conditional expectation) has the following property $\mathbb{E}[f(X)]=f(\mathbb{E[X]})$}:
\begin{align}
\label{eq:0}
\phi_i(f,x) &= \frac{1}{M!} \sum_R \mathbb{E}[f(x) \mid x_{S^R \cup i}] - \mathbb{E}[f(x) \mid x_{S^R}],\\
&= \beta \frac{1}{M!} \sum_R \mathbb{E}[x \mid x_{S^R \cup i}] - \mathbb{E}[x \mid x_{S^R}].
\end{align}

Estimating this conditional expectation is hard in general, so we assume the inputs $x\sim \mathcal{N}(\mu,\Sigma)$ are multivariate normal.  Then, denote the projection matrix that selects a set $S$ as $P_S\in (0,1)^{|S|\times |N|}$ (therefore, $P_Sx \in \mathbb{R}^{|S|\times 1}$ returns the features from $x$ in $S$), then $\mathbb{E}[x \mid x_S]$ is\footnote{In words, the conditional expectation for a normal distributed random variable is known to be $P_{\bar{S}} \mu + P_{\bar{S}} \Sigma P_S^T (P_S \Sigma P_S^T)^{-1} ( P_S x - P_S \mu)$; however, this gives us a vector in $\mathbb{R}^{|S|}$.  Since $f(x)$ expects an input in $\mathbb{R}^{|N|}$, we project the conditional expectation into $\mathbb{R}^{|N|}$ by multiplying by $P_{\bar{S}}$, where $\bar{S}\equiv N\setminus S$.  The resulting vector has all of the features not in $S$ set to zero.  These features can simply be set to their known values since we are conditioning on them, hence the addition of $x P_S^T P_S$.}:
\begin{align}
\overbrace{[P_{\bar{S}} \mu + P_{\bar{S}} \Sigma P_S^T (P_S \Sigma P_S^T)^{-1} ( P_S x - P_S \mu)]}^{\text{Conditional expectation for }x_i\in S} \underbrace{P_{\bar{S}}}_{\mathclap{\text{Project to }\mathbb{R}^{|N|}}} +
\overbrace{x P_S^T P_S}^{\mathclap{\text{Zero }x_i\notin S}}. \label{eq:1}
\end{align}

At this point, we have a natural solution to obtain the conditional expectation Shapley value for a single sample.  If we compute (\ref{eq:1}) for all sets $S^R$, we can use the combinations definition of Shapley values ($\sum_{S\in N\setminus i} W(S,N) (\mathbb{E}[x|x_{S\cup i}]-\mathbb{E}[x|x_{S}])$) to compute the Shapley value exactly.  

\textit{Computational complexity:} Each term in the summation requires a matrix multiplication/inversion which is $O(n^3)$ complexity in the size of the matrix.  Since we do this for all possible subsets, the computational complexity to obtain $\phi_i(f,x)$ is $O(|N|^3 2^{|N|-1})$.  To obtain $\phi_i(f,x)\forall i$, re-running this algorithm would result in a complexity of $O(|N|^4 2^{|N|-1})$.  Alternatively, if we re-use terms in the summation we get a complexity of $O(|N|^3 2^{|N|})$.  

Finally, to obtain $\phi_i(f,x)\forall i$ for $M$ samples, we have to incur this exponential cost $M$ times for each explanation.  Instead, we can isolate the exponential computation to a matrix that does not depend on $x$ itself.  This implies that if we can incur an exponential cost once, we can explain all samples in low order polynomial time.  To do so, we can factor (\ref{eq:1}) to get:
\begin{align}
\label{eq:2}
\mathbb{E}[x \mid x_S] = [Q_{\bar{S}} - U_S] \mu + [Q_S + U_S] x,
\end{align}
where $U_S = P_{\bar{S}}^T P_{\bar{S}} \Sigma P_S^T (P_S \Sigma P_S^T)^{-1} P_S$ and $Q_S = P_S^T P_S$.  Then, if we use equation (\ref{eq:2}) to revisit (\ref{eq:0}), we get: 
\begin{align}
\phi_i(f,x) = \beta T^{(\mu)} \mu + &\beta T^{(x)} x,
\end{align}
where $T^{(\mu)}=\frac{1}{M!} \sum_R  ([Q_{\bar{S}^R \cup i} - U_{S^R \cup i}] - [Q_{\bar{S}^R} - U_{S^R}])$ and $T^{(x)}=\frac{1}{M!} \sum_R ([Q_{S^R \cup i} + U_{S^R \cup i}] - [Q_{S^R} + U_{S^R}])$.  Here, we can see that computing $T^{(\mu)}$ and $T^{(x)}$ is exponential \footnote{In fact, the complexity to compute them for all features is $O(|N|^3 2^{|N|})$}, however, once we have computed $T^{(\mu)}$ and $T^{(x)}$, we can compute the Shapley value $\phi_i(f,x)$ quickly\footnote{In a few matrix multiplications and an addition.}.

Note that just as the original Shapley values have been approximated using Monte Carlo sampling, we can likewise approximate $T^{(\mu)}$ and $T^{(x)}$ by sampling from the permutations (or combinations) of feature orderings. In contrast to traditional sampling approaches which approximate the summation in Equation \ref{eq:0}, approximating $T^{(\mu)}$ and $T^{(x)}$ converges much faster because we do not need to separately converge for each input feature.

% \subsection{Algorithms}

% \begin{algorithm}
% \caption{Interventional Shapley values}
% \begin{algorithmic}
% \REQUIRE $\mu,x,\beta^T\in\mathbb{R}^{|N|\times 1}$, $\Sigma\in \mathbb{R}^{|N|\times |N|}$
% \STATE $\phi \leftarrow \beta (x-\mu)$
% \end{algorithmic}
% \end{algorithm}

% \begin{algorithm}
% \caption{Observational Shapley values}
% \begin{algorithmic}
% \REQUIRE $\mu,x,\beta^T\in\mathbb{R}^{|N|\times 1}$, $\Sigma\in \mathbb{R}^{|N|\times |N|}$, $n_{samp}\in \mathbb{R}^1$
% \STATE $iter \leftarrow 0$
% \WHILE{$iter < n_{samp}$}
% \STATE $S \leftarrow randomsubset(N)$
% \STATE $\phi \leftarrow \beta (x-\mu)$
% \ENDWHILE
% \end{algorithmic}
% \end{algorithm}

\section{Effects of Correlation}

\subsection{Impact of correlation on convergence}

\begin{figure*}[!ht]
    \centering
    \includegraphics[width=0.85\linewidth]{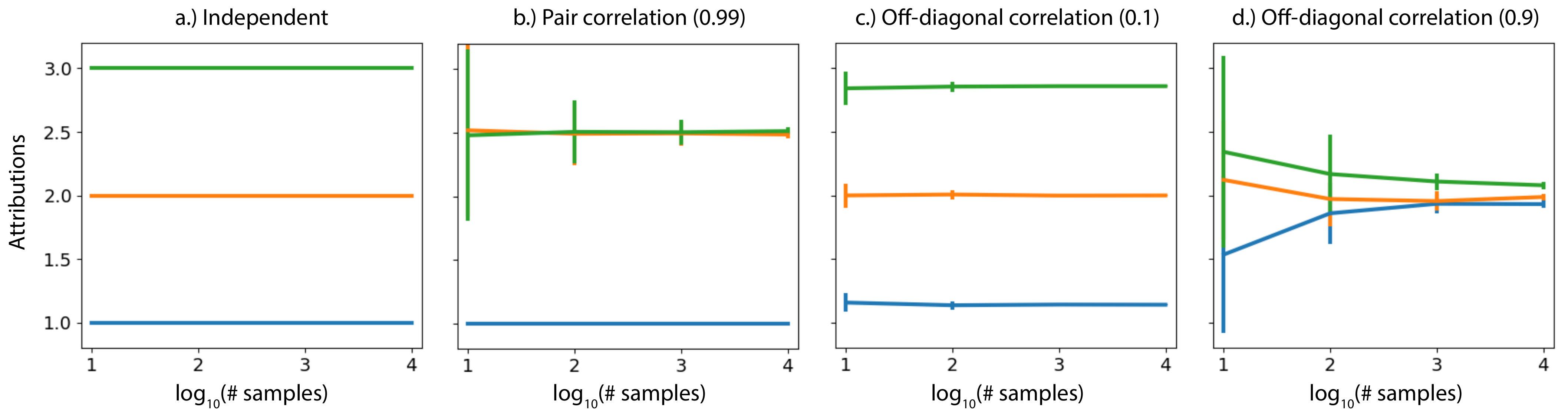}
    \caption{Convergence of correlated linear Shapley value estimates.  $x_1$'s attributions are blue, $x_2$ is orange, and $x_3$ is green.  We report the standard deviations from 20 estimates of the observational Shapley values using a fixed number of samples of combinations $S$.}
    \label{fig:convergence_correlation}
\end{figure*}

In order to build intuition about the observational Shapley values for linear models, we first examine a simulated example with a known distribution.  In the following example, the features are $x\in\mathbb{R}^3\sim \mathcal{N}(0,\Sigma)$, the model is $f(x)=1\times x_1+2\times x_2+3\times x_3$, and the sample being explained is $x^f=[1,1,1]$.

In Figure \ref{fig:convergence_correlation}, there are three cases: (1) Independent implies that the correlation is the identity matrix $\Sigma=I$, (2) Pair correlation ($\rho$) implies that $\Sigma=I$, except $\Sigma_{2,3}=\Sigma_{3,2}=\rho$, and (3) Off-diagonal correlation ($\rho$) implies that $\Sigma_{i,j}=1$ if $i=j$ and $\Sigma_{i,j}=\rho$ otherwise.  

We can observe that when features are independent, the observational Shapley value estimates are $\phi(x^f,f)=[1,2,3]$ which coincides with the interventional Shapley value estimates.  Furthermore, for data that is truly independent, there is no variance in the estimates and they converge immediately.  For other correlation patterns, we can observe two trends: 1.) correlation splits the $\beta$ as credit between correlated variables and 2.) higher levels of correlation leads to slower convergence of the observational Shapley value estimates.

\subsection{Explaining a feature not used by the model}
\label{sec:DummyFeature}

\begin{figure}[!ht]
    \centering
    \includegraphics[width=0.9\linewidth]{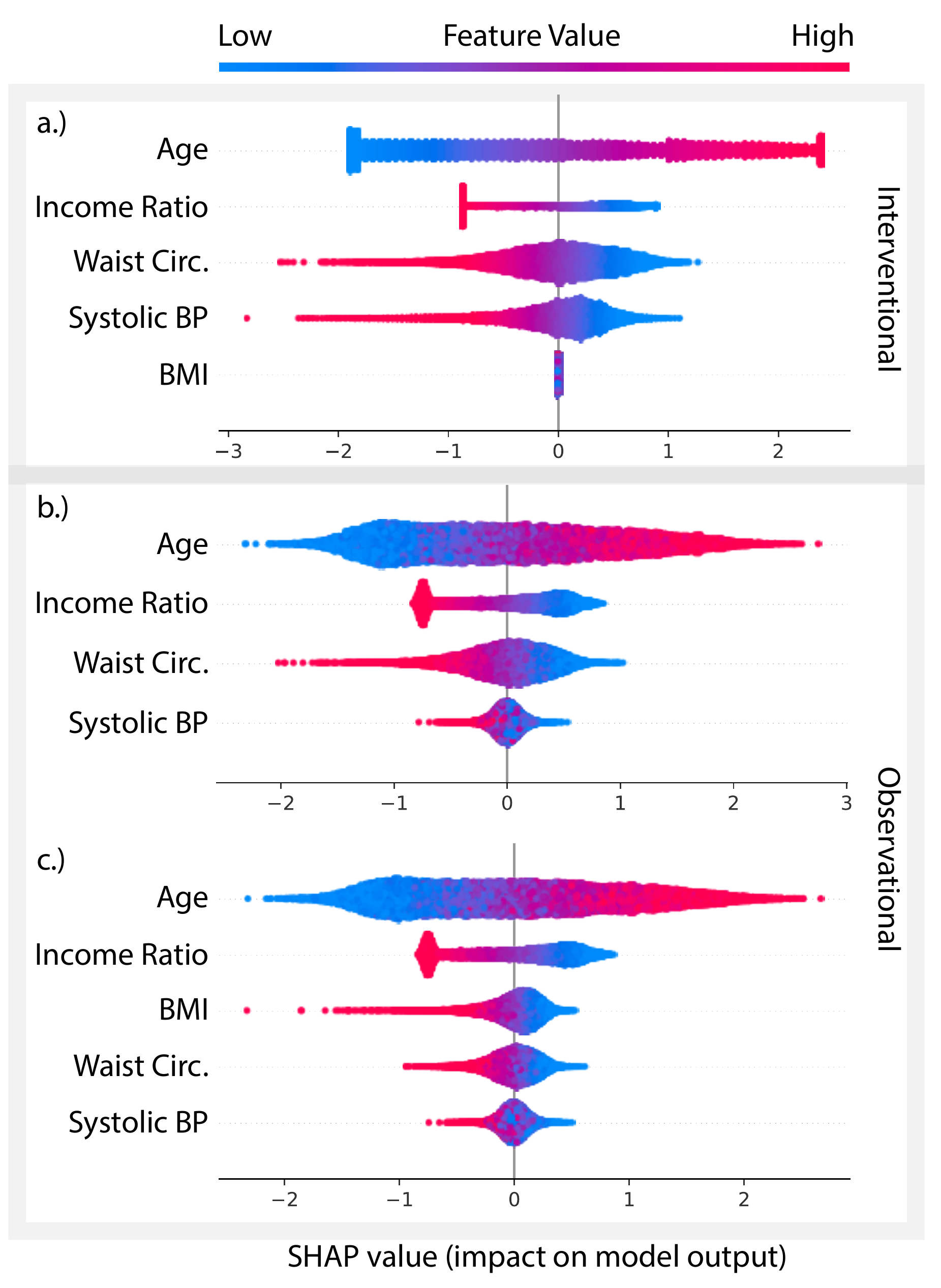}
    \caption{Interventional vs. observational SHAP values for NHANES.  In a.), we use the interventional approach whereas in b.), we use the observational approach with different sets of features.  In these summary plots, each point is an individual where the x-axis is the Shapley value, or the impact on the model's output).  The color is the value of the feature listed in the y-axis.}
    \label{fig:NHANES}
\end{figure}

In order to compare interventional/observational Shapley values on real data we utilize data from the National Health and Nutrition Examination Survey (NHANES).  In particular, we focus on the task of predicting 5-year mortality within individuals (n=25,535) from 1999-2014, where mortality status is collected in 2015\footnote{We filter out individuals with unknown mortality status.}.  Note that observational conditional Shapley values require the covariance and mean of the underlying distribution, for which we use the sampling covariance and mean.

For Figure \ref{fig:NHANES}, we use five features: \textit{Age}, \textit{Income ratio}, \textit{Systolic blood pressure}, \textit{Waist circumference}, and \textit{Body mass index (BMI)}.  In particular, we train a linear model on the first four features (excluding \textit{BMI} (test AUC: 0.772, test AP: 0.186).  Then, the interventional Shapley values give credit depending on the corresponding coefficient.  In particular, \textit{Age} positively impacts mortality prediction whereas \textit{Income ratio}, \textit{Waist circumference}, and \textit{Systolic blood pressure} all negatively impact mortality prediction.  Finally, \textit{BMI} has no importance because it is not used in the model.  

However, for the observational Shapley values, we first observe that the number of features used to explain the model impact the attributions (four features in \ref{fig:NHANES}b and five features in \ref{fig:NHANES}c).  In Figure \ref{fig:NHANES}b, we see that the relationships are similar, but slightly different to the interventional Shapley values in \ref{fig:NHANES}a due to correlation in the data.  In Figure \ref{fig:NHANES}c, we can see that when we include \textit{BMI}, the importance of the other features is relatively lower.  This implies that even though \textit{BMI} is not included in the model, the correlation between \textit{BMI} and other features makes BMI important under observational Shapley values.  Being able to explain features not in the model has implications for detecting bias.  For instance, a linear model may use correlations between features to implicitly depend on a sensitive feature that was explicitly excluded.  Observational Shapley values provide a tool to identify such bias (though we note that in this case there are other approaches to identify surrogate variables such as correlation analysis).

\section{True to the Model or True to the Data}

We now consider two examples using real world datasets and use cases that demonstrate why neither the observational nor the interventional conditional expectation are the right choice \textit{in general}, but can be chosen based on the desired application. We use these applications to argue that the choice of conditional expectation comes down to whether you want your attributions to be \textit{true to the model} or \textit{true to the data}.

\subsection{True to the Model}
\begin{figure}[!ht]
    \centering
    \includegraphics[width=0.95\linewidth]{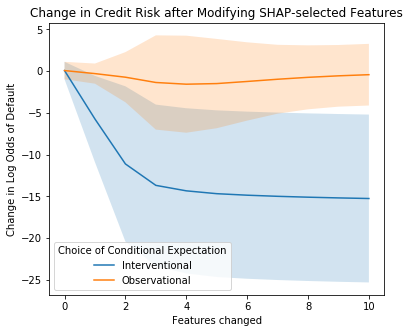}
    \caption{Modifying features according to the Interventional SHAP values helps applicants decrease their predicted log odds of default much more than Observational SHAP values. Solid line indicates mean change in log odds, while shaded region indicates standard deviation over all applicants. The wide range is expected as applicants who are very close to the mean or with very low odds of default to begin with will not be able to further decrease their odds of default by setting features to the mean.}
    \label{fig:LoansL1}
\end{figure}
We first consider the case of a bank that uses an algorithm to determine whether or not to grant loans to applicants. Applicants who have been denied loans may want an explanation for why they were denied, and to understand what they would have to change to receive a loan \cite{bhatt2020explainable}. In this case, the mechanism we want to explain is the particular model the bank uses.  In that case, we argue that we want our feature attributions to be \textit{true to the model}. Therefore, we hypothesize that the interventional conditional expectation is preferable, as it is the choice of value function that satisfies the Dummy axiom -- only features that are referenced by the model will be given importance.

To investigate this, we downloaded the LendingClub dataset\footnote{https://www.kaggle.com/wendykan/lending-club-loan-data}, a large dataset of loans issued from a peer-to-peer lending site which includes loan status and latest payment information, as well as a variety of features describing the applicants such as number of open bank accounts, age, and amount requested. We trained a logistic regression model to predict whether or not an applicant would default on their loan. We obtained feature attributions using either the observational or interventional conditional expectation.

To see which set of explanations was more useful to hypothetical applicants, we wanted to see which set of explanations helped applicants most decrease their risk of default according to the model (and consequently most increase their likelihood of being granted a loan). We therefore ranked all of the features for each applicant by their Shapley value, and allowed each applicant to ``modify their risk'' by setting that feature to the mean. We then measured the change in the model's predicted log odds of default after each feature (up to 10 features) had been mean-imputed.  For this metric, the better the explanation, the faster the predicted log odds of default will decrease. 

We find that using the interventional conditional expectation leads to significantly better results than the observational conditional expectation (\autoref{fig:LoansL1}). Intervening on the features ranked by the interventional Shapley values lead to a far greater decrease in predicted likelihood of default. In other words, the interventional Shapley values enabled interventions on individuals' features that drastically changed their predicted likelihood of receiving a loan.

When we consider the axioms fulfilled by each choice of value function, this result makes sense. As pointed out in \citet{janzing2019feature} and \citet{sundararajan2019many}, and as shown in \autoref{sec:DummyFeature}, the observational Shapley value spreads importance among correlated features that may not be explicitly used by the machine learning model. Intervening on such features \textit{will not} impact the model's output.  In contrast, the interventional Shapley value is true to the model in the sense that it gives importance to features explicitly used by the model.  For a linear model, this means the interventional approach will first change a feature $i$ where $|\beta_i (x_i - \mu_i)|$ is largest.  Compared to other features, mean imputing $x_i$ will provide the greatest change to the predicted output of a linear model. Being true to the model is the best choice for most applications of explainable AI, where the goal is to explain the model itself.

% So, when we care about a particular model, we do not want to attribute importance to features that only impact a model's output through their correlation with other features (observational Shapley values).  Instead, we want to understand which features the model is truly using (interventional Shapley values).

% Therefore, attributions calculated using the observational conditional expectation will be less useful.

\begin{figure*}[!ht]
    \centering
    \includegraphics[width=0.70\linewidth]{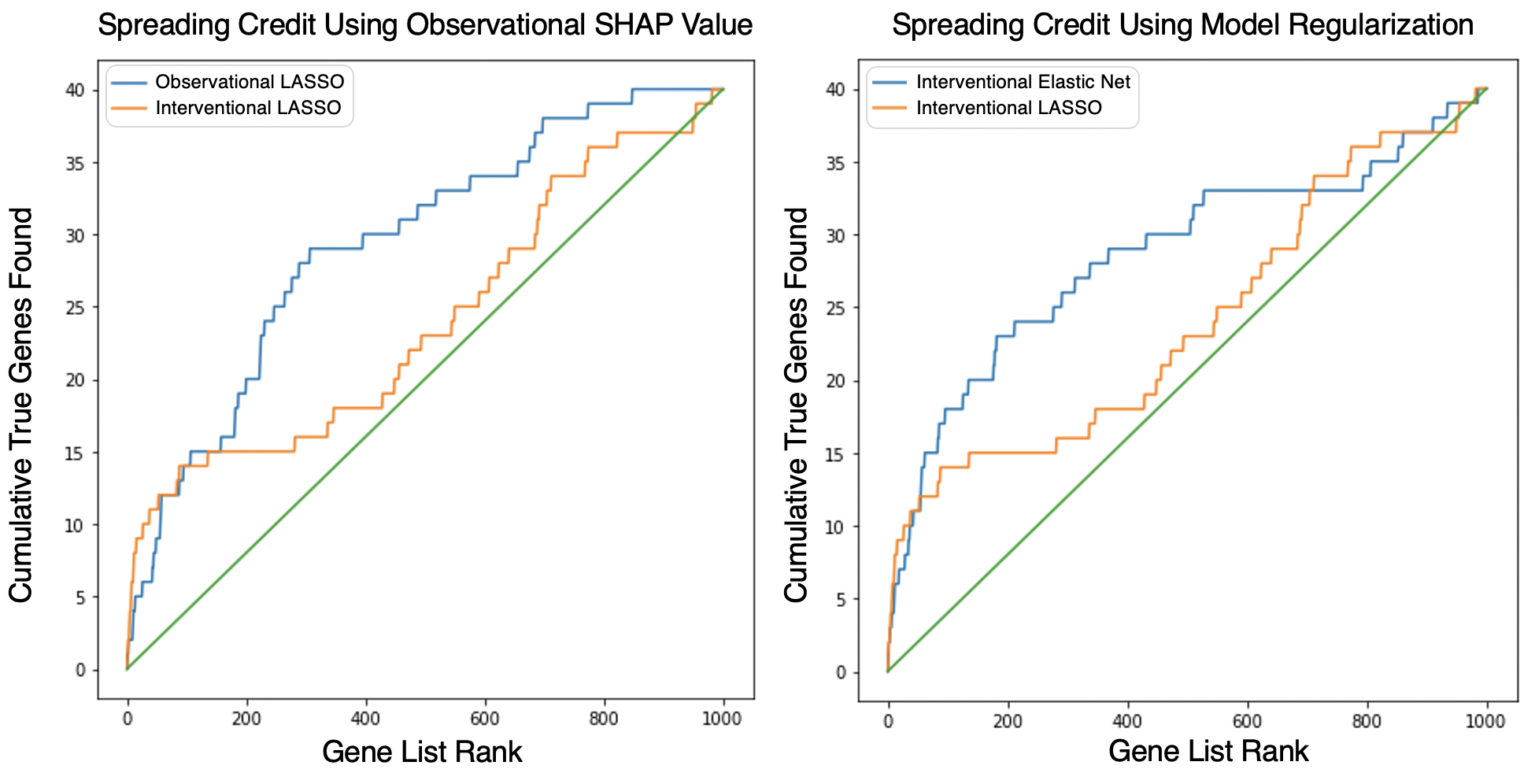}
    \caption{Left: When explaining a sparse model (Lasso regression), more true features are recovered when using the observational Shapley value to spread credit among correlated features than using the interventional Shapley value. Right: When using the interventional Shapley value, we recover more true features when the underlying model spreads credit among groups of correlated features (Elastic Net) than when the underlying model is sparse (Lasso).}
    \label{fig:rnaseqExperiment}
\end{figure*}

\subsection{True to the Data}

We now consider the complementary case where we care less about the particular machine learning model we have trained, and more about a natural mechanism in the world. We use a dataset of RNA-seq gene expression measurements in patients with acute myeloid leukemia, a blood cancer \cite{tyner2018functional}. An important problem in cancer biology is to determine which genes' expression determine a particular outcome (e.g., response to anti-cancer drugs). One common approach is to measure gene expression in a set of patient samples, measure response to drugs \textit{in vitro}, then use machine learning to model the data and examine the weights of the model \cite{zou2005regularization,lee2018machine,janizek2018explainable}.

To create an experimental setting where we have access to the ground truth, we take the real RNA-seq data and simulate a drug response label as a function of 40 randomly selected \textit{causal genes} (out of 1000 total genes). The label is defined to be the sum of the causal genes plus gaussian noise. After training a Lasso model, we explain the model for many samples using the observational and interventional Shapley values and rank the genes by their average magnitude Shapley value to get two sets of global feature importance values \cite{lundberg2020local}. We see that ranking the features according to the observational Shapley values recovers more of the true causal features at each position in the ranked list than the interventional Shapley values (\autoref{fig:rnaseqExperiment}, left). The green line in the figure represents the expected number of true genes that would be cumulatively found at each position in the ranked list if the gene list were sorted randomly. While we see that both Shapley value-based rankings outperform random rankings, the observational approach outperforms the interventional one.

This example helps to illustrate why the Dummy axiom is not necessarily a useful axiom in general. In the case of biological discovery, we do not care about the particular linear model we have trained on the data. We instead care about the true data generating process, which may be equally well-represented by a wide variety of models \cite{breiman2001statistical}. Therefore, when ranking genes for further testing, we want to spread credit among correlated features that are all informative about the outcome of interest, rather than assigning no credit to features that are not explicitly used by a single model.

While the observational Shapley value may be preferable to the interventional Shapley value when explaining a Lasso model, Elastic Net (i.e., a penalty on L1 and L2 norm of the coefficients) is actually more popular for this application \cite{zou2005regularization}. While a Lasso model may achieve high predictive performance, it will attempt to sparsely pick out features from among groups of correlated features. We re-run the same experiment, but rather than comparing the observational and interventional Shapley values applied to a Lasso model, we focus on the interventional Shapley values for (1) a Lasso model (as in the previous experiment), or (2) an Elastic Net model (\autoref{fig:rnaseqExperiment}, right). We find that by using the Elastic Net regularization penalty, the \textit{model} itself is able to spread credit among correlated features, better respecting the correlation in the data. It is worthwhile to point out here that Elastic Net models became popular for this task because it is typical practice to interpret linear models by examining the coefficient vector, which is itself an interventional style explanation (partial derivative). Since this interventional explanation does not spread credit among correlated features, it is necessary to spread the credit using modeling decisions.

We have seen that when the goal is to be \textit{true to the data}, there are two methods for spreading credit to correlated features. One is to spread credit using the observational Shapley value as a feature attribution. The other is to train a model that itself spreads credit among correlated features, in this case by training an Elastic Net regression. When we factor in the computation time for these two approaches, the choice becomes clear. Estimating the transform matrices for the observational conditional with 1000 samples took 6 hours using the CPUs on a 2018 MacBook Pro, while hyperparameter tuning and fitting an Elastic Net regression took a matter of seconds.

\section{Conclusion}

In this paper, we analyzed two approaches to explain models using the Shapley value solution concept for cooperative games. In order to compare these approaches we focus on explaining linear models and present a novel methodology for explaining linear models with correlated features.  
We analyze two different settings where either the interventional Shapley values or the observational Shapley values are preferable. In the first setting, we consider a model trained on loans data that might be used to determine which applicants obtain loans. Because applicants in this setting are ultimately interested in why the model makes a prediction, we call this case "true to the model" and show that interventional Shapley values serve to modify the model's prediction more effectively. In the second setting we consider a model trained on biological data that aims to understand an underlying causal relationship. Because this setting is focused on scientific discovery, we call this case "true to the data" and show that for a sparse model (Lasso regularized) observational Shapley values discover more of the true features. We also find that modeling decisions can achieve some of the same effects, by demonstrating that the interventional Shapley values recover more of the true features when applied to a model that itself spreads credit among correlated features than when applied to a sparse model.

Limitations and future directions: In the RNA-seq experiment we identified two solutions to identify true features: (1) Lasso regression with observational Shapley values where correlation is spread through the attribution method and (2) Elastic Net regression with interventional Shapley values where correlation is spread through the model estimation. While both approaches achieved similar efficacy, we found that the latter was far more computationally tractable. As future work, we aim to further analyze which of these approaches are preferable or even feasible for scientific discovery beyond linear models. 

Currently, the best case for feature attribution is when the features that being perturbed are independent to start with. In that case, both the observational and interventional approaches yield the same attributions. Therefore, future work that focuses on reparameterizing the model to get at the underlying independent factors is a promising approach to eliminate the \textit{true to the model} vs. \textit{true to the data} tradeoff, where we can perturb the data interventionally without generating unrealistic input values. 

% \section{Next steps, possibly after workshop?}

% Potentially the most interesting next direction is determining how difficult the trade-off between modeling the conditional distribution for the attribution is, and ``modeling the conditional'' well for the model or accounting for correlation in the model for non-linear models. Measure how hard it is for VAEAC to learn known distributions, or how well it can impute random real data distributions \cite{ivanov2018variational}. Then compare that to how hard it is to learn a model that ``respects the correlation'' in the data -- would definitely be nice to think about how to define that more precisely. Some papers that could be helpful might be the following: \cite{smilkov2017smoothgrad,ilyas2019adversarial,ghorbani2019interpretation,ross2018improving}. It would be great to connect this ``true-to-the-model'' vs. ``true-to-the-data'' debate to the whole SmoothGrad thing. Sounds super awful computationally, but it would be great in terms of ``showy'' figures to use some kind of image in-painting GAN to do SHAP values for an ImageNet classifier, compare that to SmoothGrad, and compare that to training on noisy images.

\bibliography{references}
\bibliographystyle{icml2020}

\end{document}